\documentclass[sigconf, balance=false]{acmart}
\usepackage[inline, shortlabels]{enumitem}
\usepackage{subcaption}

\begin{document}

\title{ScaleFold: Reducing AlphaFold Initial Training Time to 10 Hours}

\author{Feiwen Zhu}
\authornote{Equal contributions.}
\orcid{0009-0005-4813-0685}
\affiliation{%
  \institution{NVIDIA}%
  \city{Shanghai}%
  \country{China}%
}
\email{mzhu@nvidia.com}

\author{Arkadiusz Nowaczynski}
\authornotemark[1]
\orcid{0000-0002-3351-9584}
\affiliation{%
  \institution{NVIDIA}%
  \city{Warsaw}%
  \country{Poland}%
}
\email{anowaczynski@nvidia.com}

\author{Rundong Li}
\orcid{0009-0006-6796-210X}
\affiliation{%
  \institution{NVIDIA}%
  \city{Shanghai}%
  \country{China}%
}
\email{davidli@nvidia.com}

\author{Jie Xin}
\orcid{0009-0002-2344-4811}
\affiliation{%
  \institution{NVIDIA}%
  \city{Shanghai}%
  \country{China}%
}
\email{jxin@nvidia.com}

\author{Yifei Song}
\orcid{0009-0004-8191-3297}
\affiliation{%
  \institution{NVIDIA}%
  \city{Shanghai}%
  \country{China}%
}
\email{yifeis@nvidia.com}

\author{Michal Marcinkiewicz}
\orcid{0000-0002-1316-3293}
\affiliation{%
  \institution{NVIDIA}%
  \city{Warsaw}%
  \country{Poland}%
}
\email{michalm@nvidia.com}

\author{Sukru Burc Eryilmaz}
\orcid{0000-0002-6504-0121}
\affiliation{%
  \institution{NVIDIA}%
  \city{Santa Clara}%
  \country{United States}%
}
\email{seryilmaz@nvidia.com}

\author{Jun Yang}
\orcid{0009-0008-3059-7027}
\affiliation{%
  \institution{NVIDIA}%
  \city{Beijing}%
  \country{China}%
}
\email{juney@nvidia.com}

\author{Michael Andersch}
\orcid{0009-0004-5778-4480}
\affiliation{%
  \institution{NVIDIA}%
  \city{Berlin}%
  \country{Germany}%
}
\email{mandersch@nvidia.com}


\begin{abstract}
AlphaFold2 has been hailed as a breakthrough in protein folding. It can rapidly predict protein structures with lab-grade accuracy. However, its implementation does not include the necessary training code. OpenFold is the first trainable public reimplementation of AlphaFold. AlphaFold training procedure is prohibitively time-consuming, and gets diminishing benefits from scaling to more compute resources. In this work, we conducted a comprehensive analysis on the AlphaFold training procedure based on Openfold, identified that inefficient communications and overhead-dominated computations were the key factors that prevented the AlphaFold training from effective scaling. We introduced ScaleFold, a systematic training method that incorporated optimizations specifically for these factors. ScaleFold successfully scaled the AlphaFold training to 2080 NVIDIA H100 GPUs with high resource utilization. In the MLPerf HPC v3.0 benchmark, ScaleFold finished the OpenFold benchmark in 7.51 minutes, shown over $6\times$ speedup than the baseline. For training the AlphaFold model from scratch, ScaleFold completed the pretraining in 10 hours, a significant improvement over the seven days required by the original AlphaFold pretraining baseline.

\end{abstract}



\maketitle

\renewcommand{\shortauthors}{Feiwen Zhu, Arkadiusz Nowaczynski, et al.}

\section{Introduction}
In recent years, deep learning has proven to be effective in high performance computing. Each year, new and more accurate surrogate models are built and shown to vastly outpace physics-based simulations with sufficient accuracy to be useful. The AI-driven approach has revolutionized protein folding, with advancements seen in RoseTTAFold\cite{baek2021accurate}, AlphaFold2\cite{jumper2021highly} and other AlphaFold2-like models, such as OpenFold\cite{ahdritz2022openfold} and FastFold\cite{chen2022fast}, making protein structure-based drug discovery more accessible.

DeepMind published a detailed AlphaFold paper in Nature together with inference-only code in JAX and pretrained weights. Its accuracy is on par with the experimental methods, regarded as a solution to a 50-year-old grand challenge in biology. The AlphaFold model was built on a variant of the sequence attention mechanism~\cite{vaswani2017attention} widely adopted by other contemporary deep-learning models. The AlphaFold model was trained on over 10 million samples with 128 TPUs, took over 11 days (7 days initial training and 4 days finetuning) to converge. Such a low efficiency slows down the iterative speed of the research community.

As a result, improving the AlphaFold training performance has received increasing interest, and MLPerf HPC~\cite{farrell2021mlperf} also incorporated this challenge as a benchmark to promote broader participation\footnote{OpenFold is the first trainable public reimplementation of AlphaFold in PyTorch. Its open source release enables researchers worldwide to apply and build on this technology. OpenFold's initial training phase was selected to the MLPerf HPC\cite{farrell2021mlperf} v3.0 in early 2023. Comparing to the original OpenFold setup, the OpenFold in MLPerf HPC v3.0 is formulated as a partial convergence training, with model weights initialized from predefined checkpoint and lowered final accuracy target.}. OpenFold~\cite{ahdritz2022openfold} reproduced the AlphaFold training procedure and used BFloat16 numerical format and gradient checkpointing~\cite{chen2016training} to improve the training efficiency. DeepSpeed4Science~\cite{ds2023evo} proposed a dedicated attention kernel design that reduced memory usage. FastFold~\cite{chen2022fast} proposed Dynamic Axial Parallelism to parallelize the training with finer granularity. However, to our knowledge, none of the existing work has revealed the fundamental challenges in further accelerating the AlphaFold training.

We point out that the core challenge of improving the AlphaFold training performance is \emph{scalability}. In this study, we conducted a comprehensive analysis on the AlphaFold training procedure, and show the major causes that prevented the training from scaling to more compute resources are:
\begin{enumerate*}[1)]
  \item \emph{Communications} during the distributed training were intensive yet inefficient, largely due to communication overheads and imbalances caused by slow workers;
  \item \emph{Computation} in the training hardly saturated each worker's compute resources, due to local CPU overheads, non-parallelizable workloads, and poor kernel scalability. 
\end{enumerate*}

We proposed a collection of systematic optimizations to address these challenges. A novel non-blocking data pipeline was introduced to solve the slow-worker issue. By combining this with a series of fine-grained optimizations, which include tracing the training to CUDA Graphs to reduce overheads, the overall communication efficiency was largely improved. We identified critical computation patterns in the AlphaFold training and designed dedicated Triton kernels\cite{tillet2019triton} for each of them, fused fragmented computations throughout the AlphaFold model, and carefully tuned kernel configurations for every workload sizes and target hardware architectures. We named the training method that incorporates these optimizations \emph{ScaleFold}.

ScaleFold successfully addressed the scalability issue and scaled the AlphaFold training to 2080 NVIDIA H100 GPUs, whereas prior arts only scaled up to 512. In the MLPef HPC v3.0 benchmark, ScaledFold finished the OpenFold partial training task in 7.51 minutes, over $6\times$ faster than the benchmark baseline. For training the AlphaFold model from scratch, ScaleFold finished the pretraining (i.e., initial training phase) in 10 hours, set a new record compared to prior works.

In summary, contributions of this work are three-fold:
\begin{enumerate}[1)]
  \item We identified the key factors that prevented the AlphaFold training from scaling to more compute resources;
  \item We introduced ScaleFold, a scalable and systematic training method for the AlphaFold model;
  \item We empirically demonstrated the scalability of ScaleFold, set new records for the AlphaFold pretraining and the MLPef HPC benchmark.
\end{enumerate}

\section{Background}
In this section, we present an overview of AlphaFold and AlphaFold-like models. And then, we introduce main challenges of the AlphaFold training, including high memory consumption, massive memory-bounded kernels, suboptimal key-operation performance and limited DP(Data-Parallel) degree. Finially, we introduce FastFold's DAP(Dynamic Axial Parallelism) optimization, which was leveraged by our work.

\subsection{The AlphaFold Model} \label{sec::alpha_fold}
AlphaFold~\cite{jumper2021highly} has brought about a significant breakthrough in the field of structural biology. It is the first computational method that can regularly predict protein 3D structures from amino acid sequences with an atomic accuracy. AlphaFold introduced a novel mechanism to exchange information within the amino acid's multi-sequence alignments (MSA), and explicitly represented the 3D structure in the form of a rotation and translation for each residue of the protein. 

\begin{figure*}[htbp]
\centerline{\includegraphics[width=0.8 \textwidth]{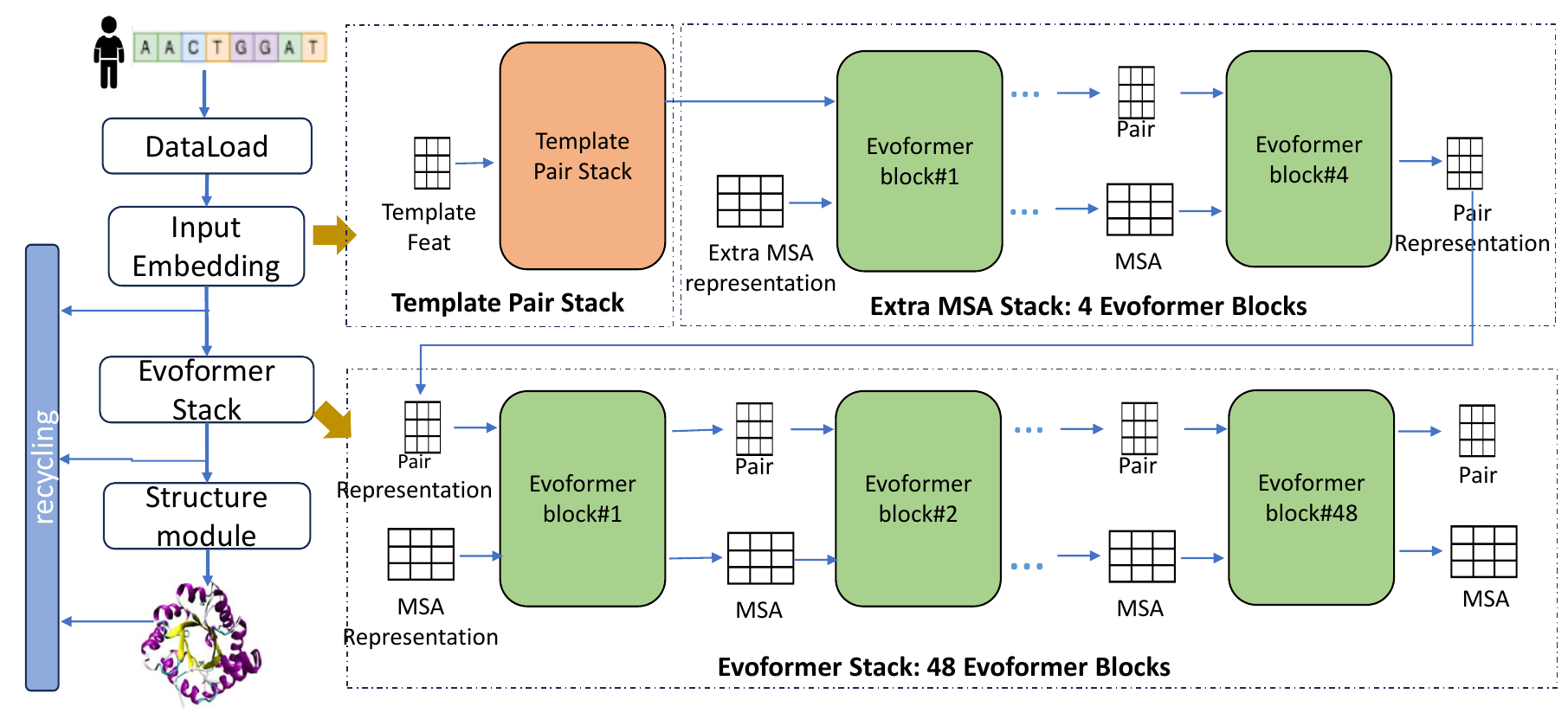}}
\caption{Structure of the AlphaFold model. Evoformer is the main building block of the AlphaFold model. In the AlphaFold model, Input Embeddings consist of Template Pair Stack, which contains 2 Evoformer blocks. Extra MSA Stack contains 4 Evoformer blocks. Evoformer stack contains 48 Evoformer blocks.}
\Description{Block diagram showing the structure of the AlphaFold model.}
\label{img::alphafold_arch}
\end{figure*}

\begin{figure*}[htbp]
\centerline{\includegraphics[width=0.9 \textwidth]{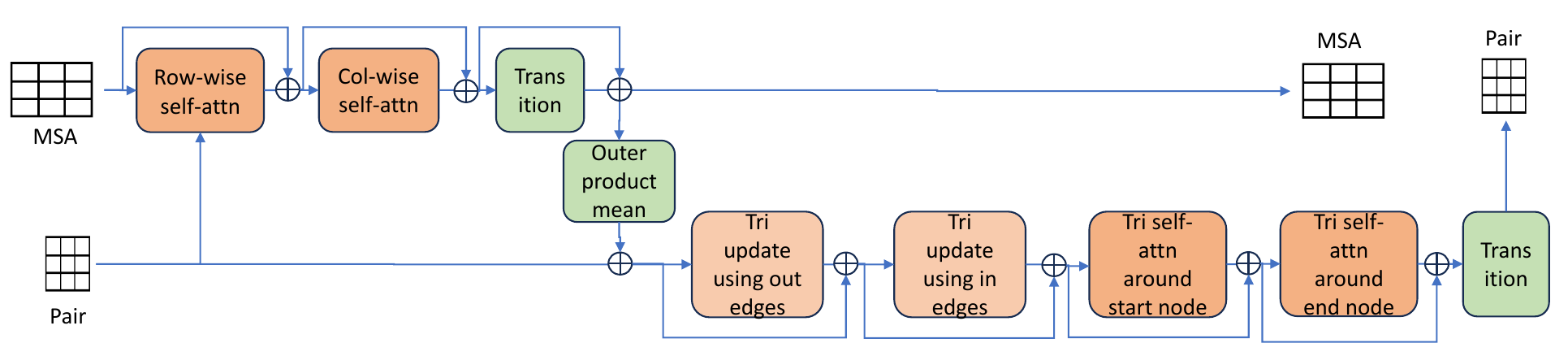}}
\caption{Structure of the Evoformer block.}
\Description{Block diagram showing the structure of the Evoformer block.}
\label{img::evoformer}
\end{figure*}

The structure of AlphaFold model is illustrated in Figure~\ref{img::alphafold_arch}. The overall architecture consists of 4 parts. \emph{Data loading} module prepares the input, MSA and template sequences, crops these sequences to a predefined length. \emph{Input Embeddings} module encodes MSA and template features of the input sequence into the initial MSA and pair representations. \emph{Evoformer Stack} module iteratively refines internal MSA and pair representations. \emph{Structure Module} outputs 3D structure for atoms in each residue. The AlphaFold training incorporates a \emph{recycling} process, enabling the continuous enhancement and fine-tuning of predicted protein structures. The core building block of the AlphaFold model is \emph{Evoformer}, which takes 72\% of each step time. Its structure is illustrated in Figure~\ref{img::evoformer}. There are 9 modules in Evoformer block, among which Row-wise self-attn module, col-wise self-attn module, triangle self-attn around starting node module and triangle self-attn around ending node module contain Multi-Head Attention.

AlphaFold was named method of the year 2021 by Science and Nature. However, its implementation does not include the necessary training code. OpenFold~\cite{ahdritz2022openfold} is a faithful reproduction of AlphaFold with a fully open-sourced training procedure and training dataset. By profiling this training procedure, the critical patterns in Evoformer are Multi-Head Attention and LayerNorm, which take $34\%$ and $14\%$ of training step time, respectively. In the rest of this work, \emph{the AlphaFold training} refers to the procedure reproduced by OpenFold.

\subsection{Challenges of the AlphaFold Training} \label{sec::alpha_fold_training}
Given the significance of the AlphaFold model, the AlphaFold training has been incorporated in the MLPerf HPC v3.0 benchmark~\cite{farrell2021mlperf}. This training procedure is highly challenging in many perspectives:

\paragraph{High Memory Consumption}
The AlphaFold model has only 97M parameters but the volume of intermediate activations during training is enormous. This is attributed to the unique attention mechanism of Evoformer, which consumes $O(n^3)$ memories for each call (see \S~1.6 of the supplementary to \cite{jumper2021highly}), significantly higher than the $O(n^2)$ memory consumption than the normal Transformer~\cite{vaswani2017attention} based models. OpenFold used gradient checkpointing~\cite{chen2016training} to mitigate this issue, yet at the cost of sacrificing the training speed.

\paragraph{Massive Memory-Bounded Kernels}
Each step of the AlphaFold training launches over 150,000 operators. Profiling results of these kernels are listed in Table~\ref{tab::kernl_breakdown}. In this table, matrix-matrix multiplications and convolutions are categorized as math-bounded kernels. Memory copy and set are categorized as memory-operation. The rests are categorized as memory-bounded kernels. The number of calls to memory-bounded kernels far exceeds that of math-bounded kernels, and they take over $65\%$ of the training time.

\begin{table}[htp]
  \small
  \centering
  \caption{Breakdown of kernels launched in the AlphaFold training. Most of these kernels are memory-bounded.}
  \label{tab::kernl_breakdown}
  \begin{tabular}{lrr}
    \toprule
    Kernel Type & Runtime (\%) & \#Calls \\
    \midrule
    CPU Overhead & 9.10 & -- \\
    Math-bounded & 24.06 & 18,147 \\
    Memory-bounded & 65.03 & 97,749 \\ 
    Memory-operation & 1.82 & 34,991 \\ 
    \bottomrule
  \end{tabular}
\end{table}

\paragraph{Suboptimal Key-Operation Performance}
In the AlphaFold training, Multi-Head Attention (MHA) and LayerNormalization (LN) are the two major performance critical operations, each of them takes $34\%$ and $14\%$ of the total training time, respectively. However, after carefully profiling the OpenFold training implementation, we found that MHA only reached $26\%$ of the theoretical performance, and LN only reached $10\%$. In addition, training routines such as optimizer update, SWA(stochastic weight average) and parameter gradient clipping were far from optimal. Weight Update takes $6\%$ of total training time and current implementation achieves $10\%$ theoretical performance. SWA takes $6\%$ of total training time and achieves less than $5\%$ of theoretical performance. gradient clipping takes $3\%$ of total training time and achieve less than $1\%$ of theoretical performance.

\paragraph{Limited Data-Parallel (DP) Degree}
To reduce training time, multi-GPU training is naturally considered. DP is the most widely used strategy to scale training to multiple workers by splitting along the batch dimension of the training samples. However, the degree of parallelism is limited by global batch size. It has been shown \cite{ahdritz2022openfold} that the training batch size of AlphaFold cannot exceed 256, otherwise it would fail to converge. This set a hard limit to the DP scaling degrees.

\subsection{Dynamic Axial Parallelism}
To scale the AlphaFold training beyond the hard limit imposed by Data-Parallel, FastFold~\cite{chen2022fast} proposed DAP(Dynamic Axial Parallelism). DAP is a model parallelism strategy, which was designed to exploit unique properties of AlphaFold2 model architecture. DAP offers several advantages over Tensor Parallelism (TP), including better scalability, lower communication volume, less memory consumption, and more opportunities for communication optimization. DAP splits intermediate activations and associated computations of a single training sample along a non-reductive axis, introducing another layer of parallelism under DP. DAP-N means N GPUs cooperating to process one sample (local batch). With the help of DAP, FastFold can increases parallelism from 128 to 512 GPUs. However, DAP requires additional communication at both forward and backward, and its scaling efficiency is suboptimal. 

\section{Scale The AlphaFold Training} \label{sec::method}
In this study, we first comprehensively analyzed the AlphaFold training performance in various parallelism strategies and identified crucial factors which prevented the AlphaFold training from scaling to more compute resources. This part of study is elaborated in \S~\ref{sec::analysis}. We then divided above factors into two categories, one for those impact the communication scalability, and another for impact the computation scalability. Our systematic solutions to address issues in each of these categories are presented in \S~\ref{sec::comm_opt} and \S~\ref{sec::compute_opt}, respectively. Combined with other optimizations we proposed, which are elaborated in \S~\ref{sec::misc_opt}, we finally scaled the AlphaFold training to 2080 NVIDIA H100 GPUs and finished the pretraining in 10 hours. We named this training method as \emph{ScaleFold}.

\subsection{Barriers to AlphaFold's Training Scalability} \label{sec::analysis}

We observed that existing approaches for scaling the AlphaFold training was suboptimal. Applying Dynamic Axial Parallelism~\cite{chen2022fast} with DAP-2 and DAP-4 to a 128-way data-parallel training only provided $1.42\times$ and $1.57\times$ speedup, respectively. And there was no performance gain on DAP-8. Ideally, the scalability of DAP-\emph{n} would provide a $n\times$ speedup.

\begin{figure}[ht]
  \centering
  \includegraphics[width=\columnwidth]{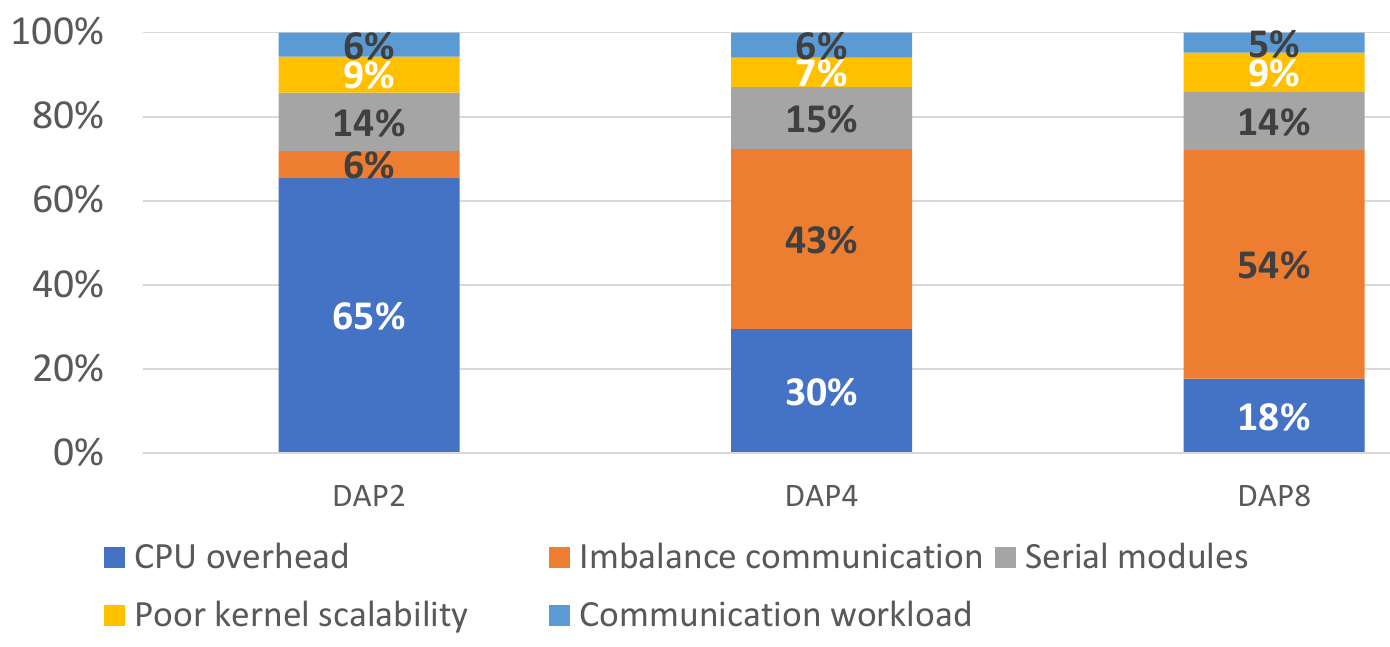}
  \caption{Breakdown of factors that prevent the AlphaFold training from achieving better scalability. Numbers indicate the relative difference between the actual time and the theoretically optimal time per training step.}
  \Description{Bar figure to demostrate factors preventing AlphaFold trining scaling.}
  \label{img::breakdown}
\end{figure}

To determine the root causes of these gaps, we ablated the contribution from each potential factor by subtracting the measured step time with the corresponding theoretically optimal time. The optimal time was calculated by assuming the mentioning factor was completely eliminated. The breakdown of analysis results on different DAP-\emph{n} are illustrated in Figure~\ref{img::breakdown}.

In the small scale of DAP-2, the CPU overhead and execution of serial modules were the major limiting factors. \textbf{CPU overhead} means the cost of launching numerous small kernels sequentially during training. This cost is largely attributed to the nature of AlphaFold model we discussed in \S~\ref{sec::alpha_fold_training}. \textbf{Serial modules} means the part of computation which cannot be parallelized by DAP. In AlphaFold, this corresponds to the data pipeline and the Structure Module, which takes $11\%$ of GPU time in total per training step.

\begin{figure}[ht]
  \centering
  \includegraphics[width=\columnwidth]{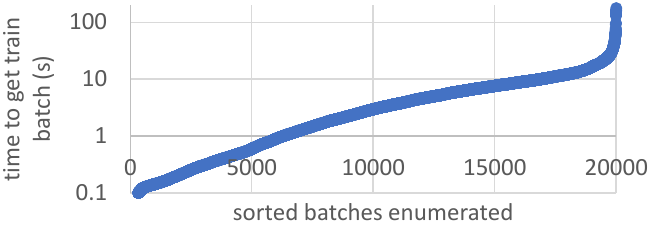}
  \caption{Sorted data batch preparation time of AlphaFold's training dataset. Depending on the data sample's initial sequence length and multi-sequence alignment size, the batch preparation time varies significantly, which could cause data pipeline blocking.}
  \Description{Plot figure to demostrate data loading time in AlphaFold training.}
  \label{img::batch_time}
\end{figure}

In larger scales of DAP-4 and DAP-8, the impact of imbalanced communication became increasingly substantial. \textbf{Imbalanced communication}. Stragglers\cite{tandon2017gradient} are a common issue in distributed training. Slow workers that fall behind the rest in reaching the synchronization point slow down the overall training progress. In AlphaFold training, this is mainly attributed to:
\begin{enumerate*}[1)]
  \item data pipeline, where $\sim10\%$ of training data batches took significantly more time to process thus blocked the training pipeline, as shown in Figure~\ref{img::batch_time}; and
  \item background processes in the cluster environment, which sporadically made CPU peaks and slowed down the corresponding workers.
\end{enumerate*}
This issue is particularly troublesome to DAP, as DAP involves numerous additional communications in both forward and backward passes. The imbalance results in extra communication latency, increasing the faster workers' NCCL kernel time. We inserted global synchronization before NCCL kernel to estimate the communication cost without imbalance issue. The difference in execution time between synchronization and no synchronization can be used to evaluate the negative impact of imbalanced communication.

In addition, we observed \textbf{poor kernel scalability} as each kernel's problem sizes reducing. DAP-\emph{n} reduces kernel workload by $n\times$. Small workload is hard to saturate GPU bandwidth, which make kernel scalability become worse. To reduce this negative impact, one potential approach is to increase the launch dimension of the kernel. We also observed \textbf{communication overhead} associated with DAP's all-gather and all-to-all communications. This is a minor factor and can be reduced by low precision.

\subsection{Reduce Communication Imbalance} \label{sec::comm_opt}
In the AlphaFold training, all local batches are cropped into the same shape. However, the time required to prepare the batches varies depending on the length of the sequences and can range across three different scales. This randomness in workload distribution leads to an imbalance. Increasing number of workers preparing batches can enhance dataload throughput, but it does not fundamentally solve the issue of workload imbalance. Moreover, background processes occupying training processes' CPU cores sporadically also results in this imbalance. We analyzed training timeline and discovered that there are always existing some CPU cores reaching 100\% utilization, which slow down the training processes scheduled to these CPU cores. To reduce the communication imbalance caused by accidental data pipeline blocking and cluster machine CPU peaks, we proposed a new data pipeline design that continuously fed data batches to the main training process, along with a module wrapper that captured the AlphaFold training in CUDA Graphs to make the training process more robust to machine CPU usage fluctuations.

\begin{figure}[ht]
  \centering
  \includegraphics[width=\columnwidth]{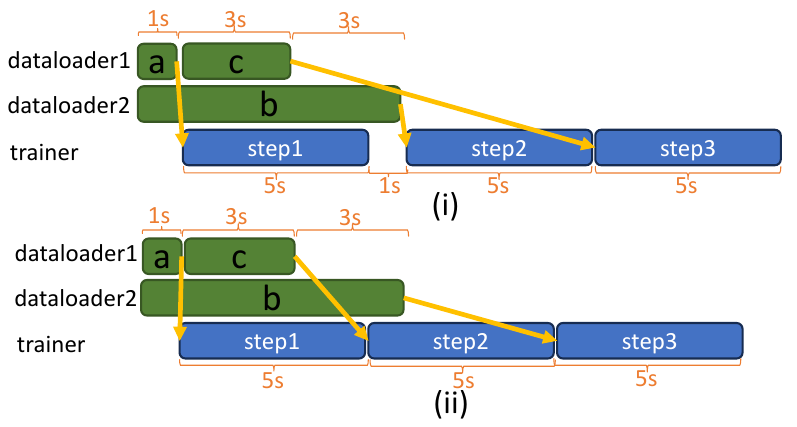}
  \caption{(i) The default PyTorch data loading pipeline vs (ii) our proposed pipeline. In PyTorch, the sampler order is enforced by its \texttt{DataLoader} even if it blocks the training: Slow batch denoted as ``b'' blocks training even though another batch ``c'' is available. In our proposed design: The batch ``c'' can be yielded before batch ``b'', which prevents imbalance and idle ranks.}
  \Description{Blocks and arrows in this figure demostrate that slow samples are sent to the training process later than faster samples.}
  \label{img::custom_dataload}
\end{figure}

\paragraph{Non-Blocking Data Pipeline}
In the AlphaFold training, the default data pipeline built on PyTorch \verb|DataLoader| generates data batches in a deterministic order. However, if the time for preparing these batches varies significantly and certain batch's preparation time exceeds the training step time a lot, that batch could be blocking the training process and make other communication participants hang. E.g., in the situation demonstrated in Figure~\ref{img::custom_dataload} (i), the slow batch \emph{b} takes 7 seconds to process, while the training step has finished at the 6th second; As a result, the training process is blocked during the whole last second.

We proposed a data pipeline that yields a batch once any of the processing batches becomes ready. This design effectively resolved the aforementioned data pipeline blocking issue. As demonstrated in Figure~\ref{img::custom_dataload} (ii), in the same situation discussed in (i), when the first training step finished at the 6th second, the data pipeline immediately yields the ready batch \emph{c} while leaving the slow batch \emph{b} being processed, so the training immediately proceed; When the second training step finished at the 11th second, the batch \emph{b} has been ready and the data pipeline feeds it to the next training step.

Processed data batches were sent to the training process via a priority queue, with the batches' indices as the associated priorities. This ensured the data yielding order in a \emph{best effort} extent. The overall data sample order could thus vary across different training instances. However, in our experiments, we did not observe any evidence showing this could negatively affect the training convergence.

\paragraph{CUDA Graph}
DAP reduces GPU workload, exposing more CPU overhead. CUDA Graph eliminates the need to interact with the CPU after graph capture, thus greatly improves training performance robustness against the CPU usage peaks. A typical way to use CUDA Graph is to define a scope, capture the computational graph within this scope, and execute the optimized graph. However, if the CUDA kernels within this scope are modified due to dynamic computation graph, such as in the case of recycling in the AlphaFold training, CUDA Graph needs to be recaptured. To address this, we designed a CUDA Graph cache that can capture multiple graphs for different recycling scenarios.

In addition, we anecdotally found that disabling Python garbage collection at runtime could alleviate machine CPU usage peaks and accelerate the overall training progress.

\subsection{Improve Computation Efficiency} \label{sec::compute_opt}
To reduce the massive CPU overheads and improve the kernel performance when its workload size is scaled down, we conducted both manual and automatic kernel fusions to reduce the kernel number and memory I/O. We also utilized the OpenAI Triton compiler~\cite{tillet2019triton} to search for optimal configurations for all generated kernels.

\subsubsection{Manual Operator Fusion and Optimization}
In AlphaFold, there are over 150,000 operations and most are memory-bound kernels. It is natural to think of reducing the number of kernels by fusion. Kernel fusion is a common optimization methodology, which combines adjacent operations' logic into a single kernel, reducing the intermediate data movement and associated kernel launching overheads.
LayerNorm and MHA (Multi-Head Attention) are critical patterns in Evoformer. We implemented efficient fused Triton kernels that fused LayerNorm, MHA and its previous four GEMMs. 
Due to the fragmented nature of the AlphaFold model, the ordinary training subroutines such as Adam optimization, stochastic weight average (SWA) and gradient clipping (grad-clip) together took $15\%$ of the training time. We implemented a customized kernel that fused Adam, SWA, along with other adjacent element-wise training logic, and parallelized this kernel across all trainable parameters. We also reordered the gradient norm calculation in grad-clip, such that its latency was perfectly hidden by distributed training communications.

\paragraph{LayerNormalization (LN)}
LN takes $14\%$ of step time in the AlphaFold training. AlphaFold's typical LN dimensions are small ($128$ and $256$), DAP further reduces problem sizes, preventing LN from fully utilizing GPU resources. We implemented a customized LN kernel to increase GPU utilization:
\begin{enumerate*}[1)]
  \item In the forward pass, we allowed each CUDA thread block to process multiple input rows;
  \item The normalization statistics were computed in a single pass, instead of using expensive iterative methods;
  \item In the backward pass, weight and bias gradients were computed by a two-step reduction, where at the first step each CUDA thread block reduced a sub-region of upstream gradients to an intermediate buffer, then at the second step each column of this buffer was reduced to obtain the final weight and bias gradients.
\end{enumerate*}
This design effectively avoided expensive atomic operations.

\begin{figure}[ht]
  \centering
  \includegraphics[width=\columnwidth]{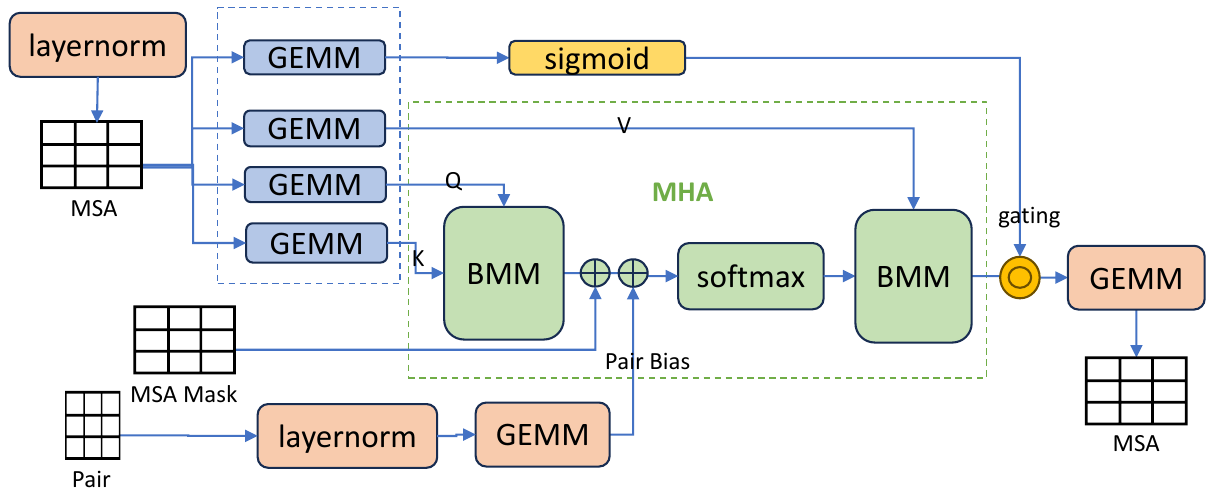}
  \caption{Structure of MSARowAttentionWithPairBias module. Its main structure is MHA.}
  \Description{Block diagram showing the structure of a row-wise attention model.}
  \label{img::row_att}
\end{figure}

\paragraph{Multi Head Attention (MHA)}
MHA takes $34\%$ of step time in the AlphaFold training. AlphaFold uses a special variant of MHA, where a \emph{pair bias} term is added to the logits matrix before the softmax operation, as shown in the dashed green box in Figure~\ref{img::row_att}. This makes integrating existing optimized MHA implementations such as FlashAttention~\cite{dao2022flashattention} inapplicable. We implemented a customized kernel based on FlashAttention~\cite{dao2022flashattention} to fuse all operations in MHA and harvested a considerable speedup.

\paragraph{GEMM Batching}
In most AlphaFold model's building blocks, the matrix-matrix multiplications (GEMMs, the dashed blue box in Figure~\ref{img::row_att}) prior to MHA do not fully leverage the potential parallelism. Four linear layers have no dependency on each other. We bundled these linear layers into batch operations to improve the degree of parallelism. 

\paragraph{Adam and SWA Optimization}. The training process incorporates the Stochastic Weight Average (SWA) to improve its convergence. However, current SWA implementation requires numerous small CUDA kernel launches, resulting in significant overheads. As SWA follows immediately after Adam optimizer, and both consist of elemwise operations, we fused Adam and SWA, along with other adjacent miscellaneous elemwise operations, into a single CUDA kernel. Each CUDA thread block handles a contiguous sub-region of elements to improve data locality, and intermediate values between Adam and SWA math are stored in GPU register files to avoid costly GPU memory operations. Additionally, we packed all parameter and optimizer state data pointers into a buffer and passed it to the fused CUDA kernel, allowing a single call to access all the elements required by Adam and SWA.

\paragraph{Gradient Clipping Optimization}. Gradient clipping typically involves three steps: 1) concatenate all parameter gradients into a vector to compute the norm of the gradient vector. 2) If the computed norm exceeds the predefined threshold, then scale down the gradients such that they remain within the specified threshold. 3) use the modified gradients for the subsequent adam optimizer.
This process can be computationally expensive in training because there are over four thousand gradient tensors at each training step. The concatenation and scaling operation each launches numerous CUDA kernels for every gradient tensors, causing significant overhead. Pytorch created a gradient buffers for distributed training, which can be reused by gradient clipping to avoid concatenating overhead. Before the distributed training, PyTorch automatically packs gradient tensors into a small number of gradient buffers. The norm of gradient can be calculated from these buffers, effectively reducing the kernel launch from thousands to tens. In addition, since the collective communications during the distributed training are performed against these buffers, the communication time perfectly hides the computation latency of the gradient clipping.

\subsubsection{Automatic Fusion and Tuning}
We do not have enough bandwidth to manually fuse all patterns. To further harvest the remaining optimization opportunities, we exploited the fusion ability provided by the \verb|torch.compile| compilation stack in PyTorch-2.0. We used it to automatically capture and fuse the fragmented operations throughout the AlphaFold model, significantly accelerated serial modules such as the Structure Module. We also found that \verb|torch.compile| did not always generate the most efficient kernels, so we controlled the compilation scope according to the target GPU architecture.

In our customized MHA and LN kernels, the OpenAI Triton compiler's auto tuning ability was exploited to search for the optimal hyper-parameters for all workload sizes that appear and target GPU architectures. The search space spanned a set of predefined tiling sizes and kernel launching dimensions. We found this to be particularly useful when workload sizes were scaled down by DAP.

\subsection{Other Optimizations} \label{sec::misc_opt}
\emph{\textbf{Low Precision}}
The MLPerf reference model only supports full precision. AMP with autocasting to fp16 converges, but it's only slightly faster than TF32 due to casting overhead and modules requiring full precision. Naive fp16 results in NaNs. We added full bfloat16 support to ScaleFold, which achieved 1.24X speedup.
\emph{\textbf{Asynchronous Evaluation and Caching Evaluation Dataset}}
As we continuously optimize step time, the proportion of evaluation time to the total training time continues to increase from 22\% to 43\%, as illustrated in Figure~\ref{fig:async_eval_perf_gain}. To eliminate the time for the evaluation, we implemented asynchronous evaluation optimization, which can offload evaluation to separated nodes and free training nodes from executing evaluation work. Evaluation time must be smaller than training time, or evaluation time would become bottleneck. so we cached all evaluation data into the CPU DRAM instead of disk to improve evaluation performance.

\section{Experiment Results}
This section presents a comprehensive evaluation of ScaleFold's performance on the NVIDIA A100 and H100 GPUs. Initially, we compared our performance to other AlphaFold2-like models. Then we assess our scale efficiency. Subsequently, we conducted a step-by-step evaluation of performance optimizations. Lastly, we evaluate the overall training time of ScaleFold on Eos cluster\footnote{NVIDIA Eos supercomputer consists of 10,752 NVIDIA H100 Tensor Core GPUs and NVIDIA Quantum-2 InfiniBand networking.}. We used CUDA version 12.2 and PyTorch NVIDIA Release 23.09 in our experiments. Each MPI task is bound to an individual GPU and 28 hyper-threading CPU cores to fully exploit CPU-GPU affinity. 8 MPI tasks are bound to a node. And all the experiments are conducted on the OpenFold dataset.

\subsection{Step Time Evaluation}
\begin{figure}[htbp]
\centerline{\includegraphics[width=0.47 \textwidth]{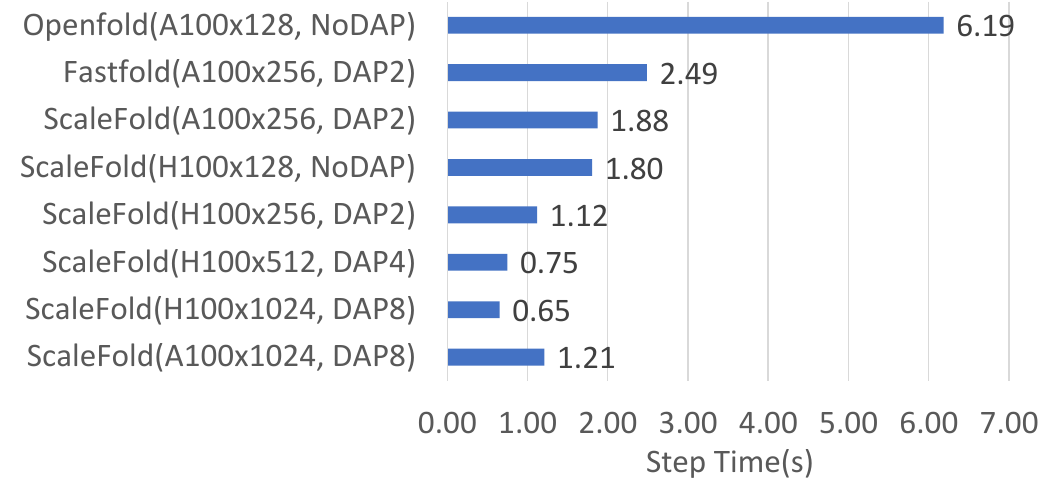}}
\caption{The step time of SacleFold on different DAP-\emph{n} compared to public OpenFold and FastFold (OpenFold and FastFold numbers come from FastFold~\cite{chen2022fast}. Batch size 128.)}
\label{fig:training_step_time}
\end{figure}

We compared the step time of our implementation to public OpenFold and FastFold. The results, as presented in Figure~\ref{fig:training_step_time}, demonstrating that the training performance of ScaleFold outperforms others. Public OpenFold doesn't support DAP and its step time on A100 is 6.19s, and the step time of FastFold DAP-2 is 2.49s on A100, while our DAP-2's step time on A100 is 1.88s. On H100, the step time of ScaleFold DAP-1 (NoDAP), DAP-2, DAP-4 and DAP-8 are 1.80s, 1.12s, 0.75s and 0.65s, respectively. So comparing to DAP-1, the speedup of DAP-2, DAP-4 and DAP-8 are 1.6X, 2.4X and 2.77X. 

\begin{figure*}[htbp]
\centerline{\includegraphics[width=0.8\textwidth]{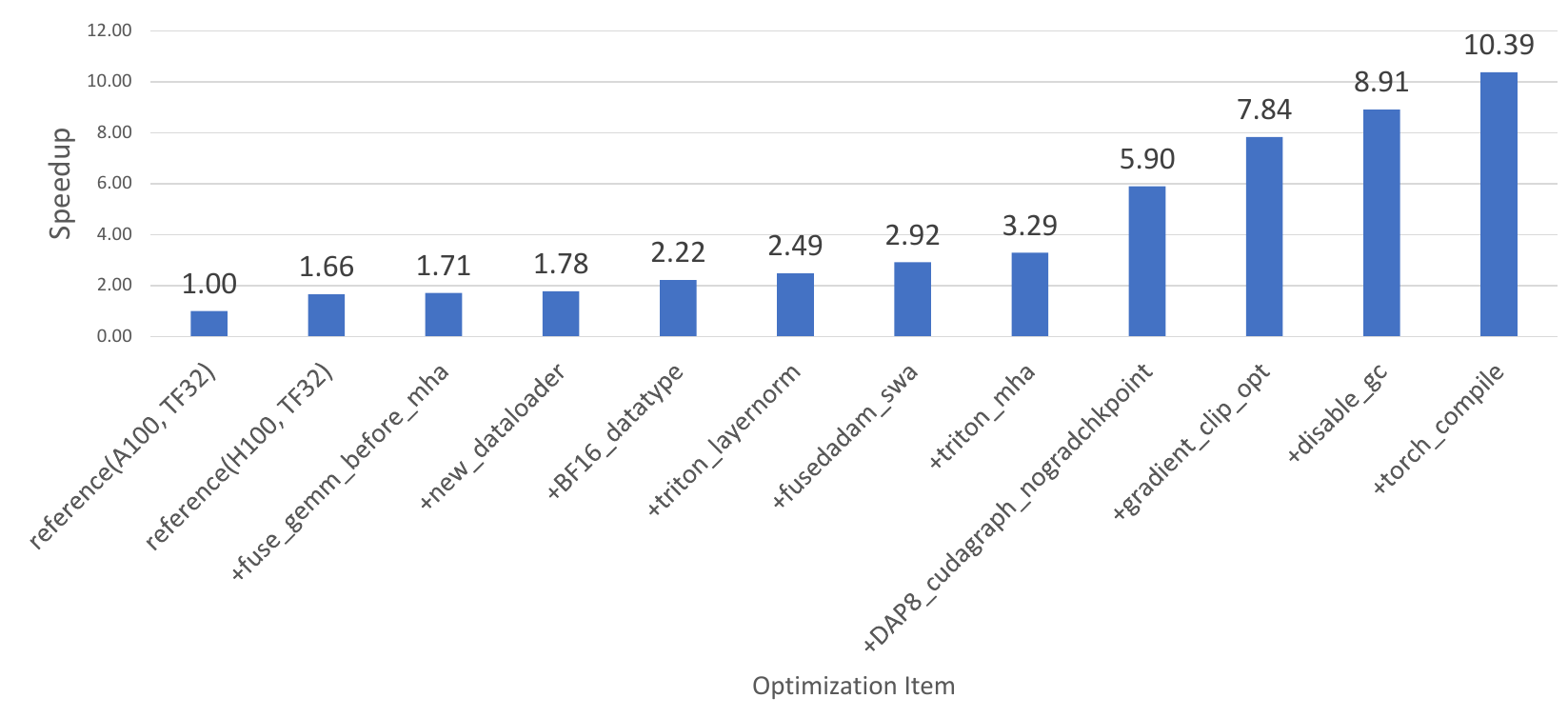}}
\caption{Step-by-step step time improvement on A100 and H100.}
\label{img::step_by_step_improve}
\end{figure*}

We conducted a comprehensive evaluation of training performance of ScaleFold on NVIDIA A100 and H100 GPU step-by-step (see Figure~\ref{img::step_by_step_improve}). Reference model requires 6.76s per step on A100, while on H100 the step time is reduced to 4.07s, which is 1.66X speedup. We observed that batching GEMM before MHA provides 1.03X speedup. Additionally, custom dataloader increased the speedup from 1.71X to 1.78X. Its speedup looks not significant, just because dataload optimization was added relatively early. As the training time continued to be optimized, the importance of dataload optimization becomes increasingly high. OpenFold is a memory bound workload, applying bfloat16 were effective in reducing memory workload, which achieved 1.24X speedup. We implemented 3 Triton kernels, MHA, LayerNorm, and FusedAdam combined with SWA. Applying these three kernels achieved 1.12X, 1.13X, and 1.17X speedup. Applying DAP reduced the pressure of memory and allowed for disabling gradient checkpointing, which eliminated re-computation in backward. CudaGraph is not beneficial for DAP-1 since DAP-1's CPU utilization is not high, but CudaGraph can be advantageous for DAP-2, DAP-4, and DAP-8. After applying DAP-8, CudaGraph and disable gradient checkpointing, we got 1.79X speedup. Without CudaGraph, DAP-8 with disabled gradient checkpointing only achieved 1.52X speedup, which was lower than DAP-4. Disabling garbage collection eliminated CPU overhead, which provided extra 1.13X speedup. Torch.compile were effective in reducing the computation and memory footprint, which provided 1.17X speedup. Finally, ScaleFold demonstrated an increased speedup of $\sim$6.2X in training step time comparing to reference model on NVIDIA H100.



\subsection{Time To Train Evaluation}
\begin{figure}[htbp]
\centerline{\includegraphics[width=0.47 \textwidth]{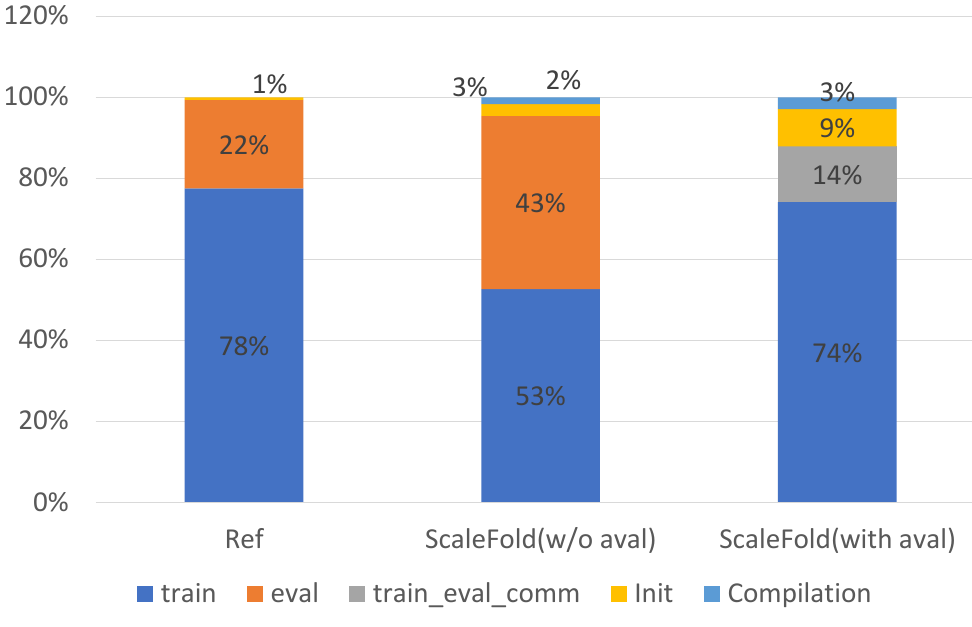}}
\caption{Breakdown Time to Train. }
\label{fig:async_eval_perf_gain}
\end{figure}

\begin{figure}[htbp]
\centerline{\includegraphics[width=0.47 \textwidth]{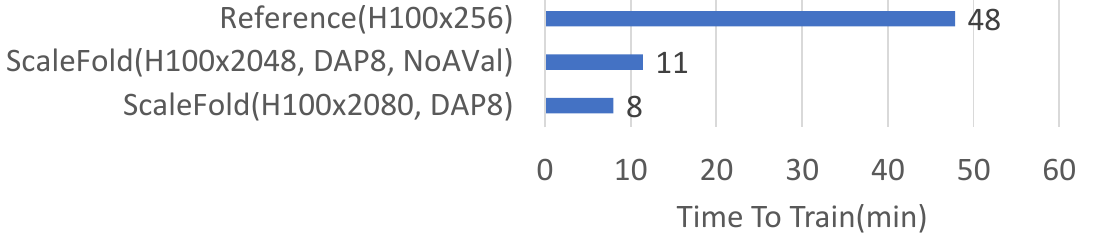}}
\caption{Time to tran from MLPerf checkpoint. Batch size 256. Reference used 256 NVIDIA H100 GPUs. ScaleFold used DAP-8.}
\label{ttt_from_chkpoint}
\end{figure}

We evaluated ScaleFold using MLPerf HPC 3.0 benchmark setting on Eos. Our largest scale was extended to 2080 NVIDIA H100 GPUs, which 2048 GPUs are used for training, and rests are used for evaluation. At largest scale, the time to train of ScaleFold was reduced to 8 minutes, including $\sim$2 minutes initialization and compilation overhead, as Figure~\ref{fig:async_eval_perf_gain}. ScaleFold is 6X faster than the reference model, as Figure~\ref{ttt_from_chkpoint}. Without asynchronous evaluation optimization and 2048 NVIDIA H100 GPUs for both training and evaluation, the train to time increased to about 11 minutes.
We also trained ScaleFold from scratch. Firstly, we used global batch size 128 to train first 5000 steps on 1056 NVIDIA H100 GPUs, with 32 of them being used for evaluation. Training metric avg\_lddt\_ca must exceed 0.8 before first 5000 training steps. Then, we used global batch size 256 and disable Triton mha kernel to train the rest steps on 2080 NVIDIA H100 GPUs (with 32 of them being used for evaluation). The whole AlphaFold pretraining requires 50000 $\sim$ 60000 steps to reach 0.9 avg\_lddt\_ca, which takes < 10 hours.


\begin{figure}[htbp]
  \centerline{\includegraphics[width=0.47 \textwidth]{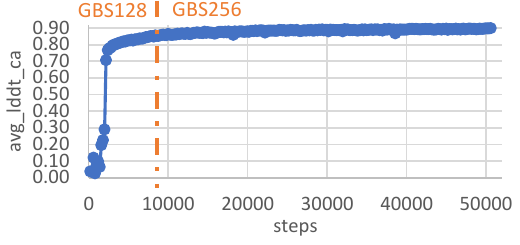}}
  \caption{AlphaFold Pretraining(initial training) from scratch. First 5000 steps uses global batch size 128, and then switch to global batch size 256.}
  \label{convergence}
  \end{figure}
\section{Conclusion}
In this work, we identified that inefficient communications and overhead-dominated computations were the two major factors that prevented the AlphaFold training from scaling to more compute resources. We introduced ScaleFold, a systematic training method for the AlphaFold model, that incorporated solutions specifically for these challenges. 
The main optimizations of ScaleFold are: 1) employed FastFold's DAP, which allows for scaling up GPU number; 2) implemented Non-Blocking Data Pipeline that eliminates negative impact of highly unequal access time to training batches; 3) enabled CUDA Graph to eliminate CPU overhead; 4) implemented efficient Triton kernels for three critical patterns, Multi-Head Attention, LayerNorm and FusedAdam combined with SWA; 5) applied torch compiler to fuse memory bound operations automatically; 6) batched GEMMs and eliminated the overhead of Gradient Clipping; 7) enabled bfloat16 training. Above optimizations reduced the step time. As step time be well-optimized, evaluation time becomes bottleneck. So we implemented 8) asynchronous evaluation to free training nodes from evaluation. Moreover, we also implemented an evaluation dataset cache to speedup evaluation time.
With these optimizations, we demonstrated its scalability and efficiency. In MLPerf HPC V3.0 OpenFold benchmark, ScaleFold's training time to convergence reduced to 7.51 minutes on 2080 NVIDIA H100 GPUs, which achieved 6X speedup of reference model. Furthermore, we trained ScaleFold from scratch and reduce initial training time from 7 days to 10 hours, as Fig.\ref{convergence}, set a new record compared to prior works. 

We hope this work can benefit the HPC and bioinformatics research community at large, by providing an effective regime to scale the deep-learning based computational methods to solve the protein folding problems. We also hope the workload profiling and optimization methodologies used in this work can shed lights on machine learning system designs and implementations.

\bibliographystyle{ACM-Reference-Format}
\bibliography{scale-fold}

\end{document}